\documentclass[11pt]{article}

\usepackage[final]{acl}

\usepackage{times}
\usepackage{latexsym}

\usepackage[T1]{fontenc}
\usepackage{CJKutf8}

\usepackage[utf8]{inputenc}

\usepackage{microtype}

\usepackage{inconsolata}

\usepackage{graphicx}
\usepackage{hyperref}
\usepackage{enumitem}
\usepackage{mdframed}
\usepackage{booktabs}
\usepackage{dblfloatfix}
\usepackage[most]{tcolorbox}
\usepackage{colortbl}
\usepackage{xcolor}

\title{%
    \textit{Liaozhai} through the Looking-Glass%
        \thanks{An allusion to footnote 44 of \citeposs{giles-1880} translation of \textit{Liaozhai}: \\[2.5ex]
            \hspace*{2em}
            \begin{minipage}{\dimexpr\linewidth-4em}
                Which will doubtless remind the reader of \textit{Alice through the Looking-glass, and what she saw there}.
            \end{minipage}
            \vspace{2ex}
        }%
    : On Paratextual Explicitation%
        \thanks{In his seminal work \textit{Seuils} \citeyearpar{genette-1987}, \citeauthor{genette-1987} defines paratexts as `thresholds' that surround and extend a text, including prefaces, glossaries, and notes like this. In translation studies, paratextual explicitation refers to clarifications and commentary appearing in such peripheral devices, rather than within the translation itself.} \\
    of Culture-Bound Terms in Machine Translation
}

\makeatletter
\let\@fnsymbol\@arabic
\makeatother

\author{%
    Sherrie Shen, \hspace{-0.2ex} Weixuan Wang, \hspace{-0.2ex} Alexandra Birch \\[1ex]
    Institute for Language, Cognition and Computation \\
    School of Informatics, University of Edinburgh \\[1ex]
    \small{
        \textbf{Correspondence:} \href{mailto:sherrie.shen@ed.ac.uk}{sherrie.shen@ed.ac.uk}
    }
}

\begin{document}
\maketitle

\begin{abstract}
The faithful transfer of contextually-embedded meaning continues to challenge contemporary machine translation (MT), particularly in the rendering of culture-bound terms—expressions or concepts rooted in specific languages or cultures, resisting direct linguistic transfer. Existing computational approaches to explicitating these terms have focused exclusively on in-text solutions, overlooking paratextual apparatus in the footnotes and endnotes employed by professional translators. In this paper, we formalize \citeposs{genette-1987} theory of paratexts from literary and translation studies to introduce the task of paratextual explicitation for MT. We const\-ruct a dataset of 560 expert-aligned paratexts from four English translations of the classical Chinese short story collection \textit{Liaozhai} and eva\-luate LLMs with and without reasoning traces on choice and content of explicitation. Experiments across intrinsic prompting and agentic retrieval methods establish the difficulty of this task, with human evaluation showing that LLM-generated paratexts improve audience comprehension, though remain considerably less eff\-ective than translator-authored ones. Beyond model performance, statistical analysis reveals that even professional translators vary widely in their use of paratexts, suggesting that cultural mediation is inherently open-ended rather than prescriptive. Our findings demonstrate the potential of paratextual explicitation in advancing MT beyond linguistic equivalence, with promising extensions to monolingual explanation and personalized adaptation.
\end{abstract}
\section{Introduction}

The intricacies in faithfully rendering meaning ac\-ross languages form a cornerstone of translation, particularly when dealing with literary devices and contextual concepts. These linguistic elements are deeply embedded within the cultural fabric of their origin, carrying connotations that can be difficult to convey into another language. Consequently, the act of translation transcends mere word-for-word substitution, requiring a nuanced understanding of the subtle interplay between language and meaning.

This challenge becomes more pronounced in the field of machine translation (MT). While modern MT systems are capable of producing fluent translations at the surface level \citep{hassan-etal-2018-achieving} and are increasingly proficient at translating idiomatic and metaphorical expressions \citep{dankers-etal-2022-transformer, karakanta-etal-2025-metaphors}, they still struggle with contextual content \citep{moghe-etal-2025-machine}. This limitation becomes especially apparent in the handling of \emph{culture-bound terms}: expressions or concepts closely associated with a particular language or culture, where the more culturally bound an item is, the greater the difficulty in translating it \citep{newmark-1988, aixela-1996}.

Translation studies has long recognized that eff\-ective cross-cultural communication requires distinct strategies tailored to a wide range of expressions, from calques and loanwords to idioms and culture-bound terms \citep{nida-1964, bassnett-1980}. Consider, for instance, the Chinese idiom `\begin{CJK*}{UTF8}{gbsn}锦上添花\end{CJK*}' (\textit{jǐn shàng tiān huā}; lit. add flowers to brocade), which means to unnecessarily embellish something already beautiful. The phrase can be translated into English as `gild the lily' since both cultures share this concept, albeit in different linguistic forms.

By contrast, culture-bound terms come with implicit nuances that often resist translation entirely. The Chinese term \begin{CJK*}{UTF8}{gbsn}江湖\end{CJK*} (\textit{jiāng hú}; lit. rivers and lakes) exemplifies this challenge:

\begin{quote}
    \textit{A semi-mythical, often romanticized social world of martial artists, wanderers, and outlaws who—despite existing outside of state authority—follow their own codes of honor and justice.}
\end{quote}

The literal translation conveys little meaningful information to a non-Chinese speaking audience. Unlike `gild the lily', no English equivalents exist for this concept, and including a full explanation within the translation would excessively disrupt narrative flow. In such cases, explicitation through external commentary offers a more elegant alternative.

\citet{genette-1987} provides a formal framework for such interventions in his theory of \emph{paratexts}: ancillary materials that accompany a primary text, mediating between the author, publisher, and reader. These liminal devices include titles, forewords, epi\-graphs, footnotes, glossaries, appendices, and other supplementary elements that, while seemingly mar\-ginal, exert a substantial influence on reader comprehension. Yet despite their pivotal role in human translation, paratexts remain entirely absent from contemporary MT—a gap that motivates our subsequent computational exploration.

\textbf{Contribution 1: Task.} In this paper, we take an initial step towards paratextual explicitation in literary MT by examining how LLMs can generate paratexts for culture-bound terms. Building upon seminal theories in literary, translation, and cultural studies, we present what is to the best of our knowledge the first formulation of this problem within computational linguistics.

\textbf{Contribution 2: Dataset.} As this task has yet to be formally studied, we introduce a dataset\footnote[3]{\url{https://github.com/sherrieshen/liaozhai}} comprising the original text and paratextual material from five published translations of the classical Chinese work \textit{Liaozhai zhiyi}. Our dataset features 560 manually-aligned paratexts across four of the translations, with expert annotations linking each paratext to the classical Chinese term it explicitates.

\textbf{Contribution 3: Evaluation.} We conduct both automatic and human evaluation of LLM-generated paratexts on our dataset, assessing how prompting and agentic strategies informed by translation theory affect performance. While these methods yield improvements of up to 5 percentage points, current LLMs continue to struggle on this task, with even the strongest model identifying fewer than 24\% of all translator-marked culture-bound terms.

\textbf{Contribution 4: Analysis.} Finally, we find that even the professional translators of our dataset systematically disagree on both choice and content of paratextual explicitation. Computational analysis reveals the inherently subjective nature of this task and provides a human upper-bound for interpreting system results.

We hope these findings will inspire further research on interdisciplinary approaches to MT and paratextual explicitation.
\section{Theoretical Foundations}

The translation of meaning across linguistic and cultural boundaries has been a subject of theoretical inquiry since antiquity. Classical scholars distinguished between \textit{verbum e verbo} (lit. word-for-word) and \textit{sensum de sensu} (lit. sense-for-sense) approaches to translation \citep{cicero-46bc, horace-19bc, st-jerome-395}, establishing a tension that would persist through centuries of scholarly discourse.

Contemporary theories have reformulated this dialectic through frameworks for equivalence and equivalent effect, most influentially articulated in \citet{nida-1964, nida-taber-1969}, and emerging alongside advances in structural linguistics \citep{chomsky-1957, catford-1965}, comparative stylistics \citep{vinay-darbelnet-1958}, and translational typology \citep[\textit{inter alia}]{newmark-1981, koller-1995}. Beyond lexical and syntactic fidelity, however, theorists increasingly recognized the importance of contextual and pragmatic factors, paving the way for later reconceptualizations of translation as a cultural practice.

\subsection{From Linguistic to Cultural Transfer}

In the 1990s, translation studies experienced the `cultural turn', a shift that aligned the field with broader intellectual movements across the humanities \citep{bassnett-lefevere-1990}. Drawing upon poststructuralism, postcolonial theory, and cultural studies, scholars began to question traditional assumptions of textual authority, cultural negotiation, and the politics of representation, with translators no longer seen as invisible agents \citep{venuti-1995} but active mediators \citep{lee-2013}.

Central to the cultural turn was recognition of translation as a form of rewriting, with translators inevitably reshaping texts to abide by literary conventions and ideological norms. Historical examples illustrate the ethical complexities: early Victorian and Edwardian translators systematically softened explicit themes of classical Greek comedies like \textit{Lysistrata} \citep{lefevere-1992}; Edward Fitzgerald's celebrated rendering of Omar Khayyam's \textit{Rubáiyát} reimagined the Persian source according to his own aesthetic ideals \citep{davis-2000}; and colonial-era translations frequently marginalized indigenous voices through assimilation into Western interpretive frameworks \citep{spivak-1978}.

These cases underscore translation as a site of cultural negotiation, where adaptation to a target context can easily blur into distortion of the source. The tension between mediation and manipulation has since motivated subsequent theoretical models, particularly those addressing the role of translation as both a cultural bridge and an ideological filter.

\subsection{Theories in Translation Studies}
\label{sec:theories_in_translation_studies}

Two theoretical frameworks emerging from the cultural turn prove particularly relevant to understanding strategies for adaptation. Polysystem theory, for one, challenges the traditional notion of texts as an isolated entity by conceptualizing translated literature as part of a dynamic system of cultural products constantly interacting within target societies \citep{even-zohar-1978, toury-1995}.

Central to polysystem theory is the idea that the position of translated literature within a given literary system fundamentally determines translational strategies \citep{munday-2016}. When occupying a primary role within the polysystem, translators tend to adopt experimental approaches that challenge existing conventions. Conversely, when relegated to a peripheral role, translations typically conform to established target-culture expectations \citep{tymoczko-2010}.

Complementing this perspective, skopos theory foregrounds the \textit{skopos}—or intended function—of the translated text within its target context \citep{reiss-vermeer-1984}. This communicative approach acknowledges that the purpose of the translated text may not always match that of the original and questions the traditional view of a translator's task in producing `equivalent effect' between two languages \citep{munday-2016}.

Together, these frameworks illuminate how translation strategies are conditioned by both the literary systems they enter and the functional purposes they fulfill. This theoretical foundation provides a conceptual basis for understanding translator use of paratexts and analyzing translations within their broader cultural contexts, as we explore in the following sections.
\section{Dataset}

\begin{table*} \small
    \centering
    \begin{tabular}{lrcccccc}
    \toprule
    \textbf{Translator(s)} & \textbf{Year} & \textbf{Profession(s)} & \textbf{Direction} & \textbf{Stories} & \textbf{Paratexts} & \textbf{Tokens} & \textbf{Types} \\
    \midrule
    \citeauthor{giles-1880} & \citeyear{giles-1880} & diplomat, professor & \textsc{fl→nl} & 164 & 657 & 28,076 & footnotes, appendices \\
    \citeauthor{lu-etal-1982} & \citeyear{lu-etal-1982} & professors & \textsc{nl→fl} & 59 & 102 & 2,210 & footnotes \\
    \citeauthor{mair-and-mair-1989} & \citeyear{mair-and-mair-1989} & sinologists & \textsc{fl→nl} & 43 & 95 & 1,695 & footnotes \\
    \citeauthor{minford-2006} & \citeyear{minford-2006} & professor, sinologist & \textsc{fl→nl} & 86 & 238 & 14,434 & endnotes, glossary \\
    \citeauthor{sondergard-2008} & \citeyear{sondergard-2008} & professor & \textsc{fl→nl} & 76 & 257 & 6,364 & footnotes \\
    \bottomrule
    \end{tabular}
    \caption{For the five English translations of \textit{Liaozhai} in our dataset, we report translator background, direction of translation (native language: NL; foreign language: FL), number of stories, number of paratexts, token count of paratexts, and type of paratexts.}
    \label{tab:dataset_statistics}
    \vspace{-2.5ex}
\end{table*}

Constructing a dataset of paratextual explicitations requires identifying a source text that can meaningfully challenge computational models. The text should demand deep cultural and contextual understanding, so that success depends on more than surface-level translation. It should exist within a long history of translation, offering diverse human-authored paratexts against which model outputs can be evaluated. Finally, it ideally takes the form of short narratives, thereby reducing confounds such as the long-range dependencies and memory limitations of current LLMs. Guided by these principles, we choose the following dataset for evaluating culture-bound paratextual explicitation.

\subsection{Source: \textit{Liaozhai zhiyi}}

We select the classical Chinese short story collection\begin{CJK*}{UTF8}{gbsn}《聊斋志异》\end{CJK*}(\textit{Liáozhāi zhìyì}, or \textit{Liaozhai} for short) as the source text for this task. Composed by Pu Songling during the Qing dynasty \citeyearpar{pu-1766}, this canonical collection of 494 stories epitomizes the \textit{zhiguai} (supernatural) literary genre.

\textit{Liaozhai} interweaves fantastical storytelling on encounters with fox spirits and ghosts through social satire and philosophical reflection, frequently using idiomatic expressions, historical allusions, and culture-bound terms \citep{yi-etal-2025}. It holds remarkable status in Chinese literature, and since its introduction to the West, has been the subject of over forty English translations \citep{jin-2021}. This yields a rich body of paratextual commentary that reflects the diverse backgrounds and interpretive styles of its various translators across the past three centuries.

\subsection{Translation Selection and Processing}

From the wide corpus of \textit{Liaozhai} translations, we filter for English editions that (1) include at least fifty stories; (2) contain substantial paratextual material; (3) are publicly available; and (4) were published at least five years prior to this study.

Of the five remaining translations, we notably decide to exclude \citet{giles-1880} from evaluation, though still release it to support future research. While Giles established the primary point of entry for \textit{Liaozhai} in the Western literary polysystem, he diverges substantially from the source material and makes culturally-outdated commentary. Due to this and reasons detailed in Appendix~\ref{appendix:giles}, we deem his translation unsuitable for comparison.

The remaining four translations form the basis of our evaluation. Each edition was OCR-processed and aligned with its corresponding classical Chinese source following \citeposs{zhang-1978} index. To comply with fair use regulations, we omit the story translations themselves and retain only paratexts and their associated metadata. A professional editor reviewed the corpus in full and identified thirty-seven typographical and factual issues—including errors in spelling, grammar, punctuation, formatting, and historical detail—all of which were corrected and documented. For experiments and analyses, we combine all terms identified by the four translators into a single, comprehensive reference set, ensuring that no culture-bound explicitation introduced by any individual translator is overlooked. Comprehensive dataset statistics, including counts for the unused \citet{giles-1880}, are detailed in Table~\ref{tab:dataset_statistics}.

\subsection{Annotation}

A professional translator manually aligned each of the 692 paratexts across the four non-Giles translations with their corresponding classical Chinese source. In instances where multiple translators explicated the same phrase, entries were consolidated, resulting in 560 unique culture-bound terms.

To facilitate downstream analysis, we classified these paratexts according to a five-part framework derived from polysystem theory:

\begin{itemize}[noitemsep]
    \item \textbf{literary} explicitations clarify stylistic devices and narrative techniques;
    \item \textbf{historical} explicitations provide information on figures, events, and periods;
    \item \textbf{cultural} explicitations address beliefs, customs, celebrations, and folklore;
    \item \textbf{social} explicitations explain social structures, hierarchies, or relationships; and
    \item \textbf{supplementary} explicitations cover any other form of contextual commentary.
\end{itemize}

\noindent All annotations are provided in our \href{https://github.com/sherrieshen/liaozhai}{released dataset}, with a representative example in Figure~\ref{fig:dataset_example} and examples of each classification in Appendix~\ref{appendix:subsystem_examples}.

\begin{figure}
    \begin{mdframed}
        \setlength{\parskip}{0.5em}
        \textbf{Classical Chinese Term:} \begin{CJK*}{UTF8}{gbsn}青山白云人\end{CJK*}
        
        \textbf{Literal translation (GPT-4o):} `person of green mountains and white clouds'
        
        \textbf{Paratextual explicitation:} \textit{Fu Yi}: A Tang dynasty scholar (555-639), who wrote his own epitaph \citep{mair-and-mair-1989}.
        
        \textbf{Subsystem}: Historical
    \end{mdframed}
    \caption{Paratext alignment and classification in the \textit{Liaozhai} dataset. While the source term appears to be a descriptive phrase, \citet{mair-and-mair-1989} connects it to historical figure Fu Yi.}
    \label{fig:dataset_example}
    \vspace{-0.5ex}
\end{figure}
\section{Experiments}

\begin{table*}[b] \small
    \centering
    \begin{tabular}{lcccccccccccc}
         \toprule
         \multicolumn{1}{c}{} & \multicolumn{6}{c}{\textbf{\textsc{Qwen3-8B} (non-thinking})} & \multicolumn{6}{c}{\textbf{\textsc{Qwen3-8B} (thinking})} \\
         \cmidrule(lr){2-7} \cmidrule(lr){8-13}
          & \textbf{TP} & \textbf{FP} & \textbf{FN} & \textbf{P} & \textbf{R} & \textbf{F1} & \textbf{TP} & \textbf{FP} & \textbf{FN} & \textbf{P} & \textbf{R} & \textbf{F1} \\
         \midrule
         \textbf{Default} & 161 & 2351 & 399 & 6.41\% & 28.75\% & 10.48\% & 201 & 1699 & 359 & 10.58\% & 35.89\% & 16.34\% \\
         \textbf{Theoretical} & \textbf{191} & \textbf{1869} & \textbf{369} & \textbf{9.27\%} & \textbf{34.11\%} & \textbf{14.58\%} & 215 & \textbf{1037} & 345 & \textbf{17.17\%} & 38.39\% & \textbf{23.73\%} \\
         \textbf{Practical} & 182 & 2071 & 378 & 8.08\% & 32.50\% & 12.94\% & \textbf{216} & 1482 & \textbf{344} & 12.72\% & \textbf{38.57\%} & 19.13\% \\
         \bottomrule
    \end{tabular}
    \caption{\textbf{Culture-bound term identification results for \textsc{Qwen3-8B} under non-thinking and thinking modes.} Results are reported as true positives (TP), false positives (FP), false negatives (FN), precision (P), recall (R), and F1. Best-performing settings within are in bold.}
    \label{tab:term_identification}
\end{table*}

We design a two-step pipeline for generating paratextual explicitations in literary MT. Given a story from \textit{Liaozhai}, the system proceeds as follows:

\begin{enumerate}[noitemsep]
    \item Identifies source terms requiring explicitation
    \item Generates a paratext from the story context
\end{enumerate}

\subsection{Culture-Bound Term Identification}
\label{sec:term_identification}

The first step identifies candidate terms from the source text requiring explicitation. For the 150 stories in our dataset containing at least one translator paratext, we prompt the model with the complete classical Chinese story and instruct it to extract expressions likely to demand additional explanation in translation. We compare three prompt variants:

\begin{itemize}
    \item \textbf{Default:} A baseline prompt that directs the model to identify terms requiring explanation when translated, with no additional framing.

    \item \textbf{Theoretical:} The baseline prompt augmented with an explicit reference to and explanation of culture-bound terms. Its design is informed by polysystem theory, as introduced in \S\ref{sec:theories_in_translation_studies}.
    
    \item \textbf{Practical:} The baseline prompt augmented with instructions on the communicative function of translation. Its design follows Skopos theory, as similarly introduced in \S\ref{sec:theories_in_translation_studies}.
\end{itemize}

\noindent Complete prompt templates are provided in Appendix~\ref{appendix:prompts_identification}. All experiments use the ChatML framework, with non-default extensions appended as additional user turns. We opt for zero-shot prompting to evaluate a model's intrinsic ability to identify culture-bound terms, deliberately avoiding few-shot exemplars to reduce bias.

Model outputs are evaluated via partial substring matching: a prediction is considered correct if and only if it contains or is contained by any translator-annotated term for the corresponding story. Performance is reported using standard information retrieval (IR) metrics: true positives, false positives, false negatives, precision, recall, and F1.

\subsection{Culture-Bound Term Explicitation}
\label{sec:term_explicitation}

For each of the 560 deduplicated terms extracted from the classical Chinese source text, we prompt the LLM to generate paratextual explicitations under conditions corresponding to the \emph{Default}, \emph{Theoretical}, and \emph{Practical} formats described in \S\ref{sec:term_identification}. Complete prompts are provided in Appendix~\ref{appendix:prompts_explicitation}.

Model outputs are evaluated against the \mbox{pool of} all translator paratext(s) for that story using \mbox{a suite} of four lexical, semantic, and LLM-based metrics: BLEU \citep{papineni-etal-2002-bleu}, ROUGE-L \citep{lin-2004-rouge}, BERTScore \citep{zhang-etal-2020}, and LLM-as-a-Judge \citep{zheng-etal-2023}. Where multiple references exist, all are included in the evaluation.

\subsubsection{Agentic Retrieval}

To mimic the way human translators consult external resources, we extend the strongest-performing model with search capabilities to scrape the web for relevant knowledge. Implemented using \href{https://www.langchain.com/langgraph}{LangGraph}, the agent executes a query given a culture-bound term and its surrounding story context.

First, it generates candidate English translations of the term and queries both Chinese (\href{https://www.baidu.com}{Baidu}) and English (\href{https://www.google.com}{Google}) search engines to retrieve information from bilingual sources. This approach enables access to long-tail knowledge in the source language, where culture-specific references are often better documented, while also capturing terminology conventions in the target language.

Returned results are appended as additional context for paratext generation. Importantly, we do not provide the agent with ground-truth translations, as identifying contextually-appropriate renderings is itself a central part of explicitation. Finally, we opt not to perform retrieval over a specific knowledge base to maintain broader applicability.

\subsection{End-to-End Evaluation}

To assess overall system performance, we conduct an integrated evaluation in which the LLM must both identify and explicitate culture-bound terms without access to gold-standard annotations. In the direct inference setting, the model executes identification and explicitation prompts sequentially. In the explicit chain-of-thought setting, the model is instructed to perform both steps in a single generation step while producing intermediate reasoning traces. This end-to-end evaluation offers a more holistic measure of system capabilities, capturing performance across the full pipeline rather than isolated subtasks.
\section{Results}

\begin{table*} \small
    \centering
    \begin{tabular}{lcccccccccc}
        \toprule
        \multicolumn{1}{c}{} & \multicolumn{4}{c}{\textbf{\textsc{Qwen3-8B} (non-thinking)}} & \multicolumn{4}{c}{\textbf{\textsc{Qwen3-8B} (thinking)}} \\
        \cmidrule(lr){2-5} \cmidrule(lr){6-9}
         & \textbf{BLEU} & \textbf{ROUGE-L} & \textbf{BERTScore} & \textbf{LLM} & \textbf{BLEU} & \textbf{ROUGE-L} & \textbf{BERTScore} & \textbf{LLM} \\
        \midrule
        \textbf{Default} & 1.40 & 12.91 & 83.37 & 64.91 & 1.31 & 15.25 & 84.61 & 68.72 \\
        \textbf{Theoretical} & 1.25 & 12.94 & 83.51 & 68.42 & 1.08 & 15.14 & 85.21 & 71.70 \\
        \textbf{Practical} & \textbf{1.57} & \textbf{13.88} & \textbf{84.36} & \textbf{69.39} & 1.62 & 16.82 & 85.56 & 72.83 \\
        \textbf{Agentic} & – & – & – & – & \textbf{2.14} & \textbf{20.59} & \textbf{86.08} & \textbf{75.69} \\
        \bottomrule
    \end{tabular}
    \caption{\textbf{Automatic evaluation of paratextual explicitation quality.} \textsc{Qwen3-8B} is evaluated under non-thinking and thinking conditions using BLEU, ROUGE-L, BERTScore, and LLM-as-a-Judge (LLM). \textit{Practical} prompting yields the strongest results without retrieval, and the agentic setup achieves the highest overall performance (in bold).}
    \label{tab:term_explicitation}
\end{table*}

\begin{table*} \small
    \centering
    \begin{tabular}{lcccccccccc}
        \toprule
        \multicolumn{1}{c}{} & \multicolumn{4}{c}{\textbf{\textsc{Qwen3-8B} (non-thinking)}} & \multicolumn{4}{c}{\textbf{\textsc{Qwen3-8B} (thinking)}} \\
        \cmidrule(lr){2-5} \cmidrule(lr){6-9}
         & \textbf{BLEU} & \textbf{ROUGE-L} & \textbf{BERTScore} & \textbf{LLM} & \textbf{BLEU} & \textbf{ROUGE-L} & \textbf{BERTScore} & \textbf{LLM} \\
        \midrule
        \textbf{Default} & \textbf{0.01} & \textbf{0.09} & 53.34 & 21.11 & 0.01 & 0.11 & 53.03 & 30.73 \\
        \textbf{Theoretical} & 0.01 & 0.08 & 55.45 & 22.28 & 0.01 & 0.10 & 55.54 & 32.89 \\
        \textbf{Practical} & 0.01 & 0.09 & \textbf{57.49} & \textbf{24.82} & \textbf{0.02} & \textbf{0.15} & \textbf{57.65} & \textbf{35.41} \\
        \bottomrule
    \end{tabular}
    \caption{\textbf{End-to-end evaluation combining term identification and explicitation.} \textsc{Qwen3-8B} performance under non-thinking and thinking conditions, assessed with BLEU, ROUGE-L, BERTScore, and LLM-as-a-Judge (LLM). While absolute scores remain low, \textit{Practical} prompting produces the most consistent gains (in bold).}
    \label{tab:end_to_end}
    \vspace{-2ex}
\end{table*}

Our experiments are structured into three stages: first, culture-bound term identification across the 150 stories in our \textit{Liaozhai} dataset (\S\ref{sec:identification_results}); second, paratextual explicitation of the 560 deduplicated translator-annotated terms (\S\ref{sec:explicitation_results}); and third, combined evaluation of term identification and explicitation in an end-to-end pipeline (\S\ref{sec:end_to_end_results}). We report results under two inference modes: \texttt{non-thinking}, in which the model generates outputs directly, and \texttt{thinking}, in which the model produces intermediate reasoning tokens before final output. This follows the terminology of \textsc{Qwen3-8B} \citep{qwen3-2025}, the model we conduct all experiments on (unless otherwise specified). Experimental setup details are provided in Appendix~\ref{appendix:experimental_setup}.

\subsection{Culture-Bound Term Identification}
\label{sec:identification_results}

To evaluate LLM performance on term identification, we test the three prompting strategies in \S\ref{sec:term_identification} against our pooled set of 560 culture-bound terms. Use of standard IR metrics allows us to assess the degree to which translation studies-informed instructions improve a model's ability to detect culturally-significant expressions.

As shown in Table~\ref{tab:term_identification}, the \emph{Theoretical} prompt consistently outperforms both \emph{Default} and \emph{Practical} prompts across inference modes. Under the non-thinking setting, it achieves the highest F1 through balanced gains in precision and recall. This improves further in the thinking setting, with precision nearly seven points higher than that of the \emph{Default} baseline. Although the \emph{Practical} prompt yields the highest recall, its lower precision results in an overall weaker F1.

These results indicate that prompts grounded in translation theory enable the model to recover culture-bound terms while minimizing noise. This aligns with the intuition that theoretical frameworks provide explicit criteria for identifying culturally-marked elements and assessing their translational relevance. Thinking mode further reduces false positives, reflecting the model's two-stage hypothesis-and-filter process.

\subsection{Culture-Bound Term Explicitation}
\label{sec:explicitation_results}

Once culture-bound terms have been identified, the next step is then to generate paratexts. We evaluate outputs using lexical (BLEU, ROUGE-L), semantic (BERTScore), and LLM-based metrics (LLM-as-a-Judge with \textsc{Qwen3-235B-A22B}).

In contrast to term identification, results in Table~\ref{tab:term_explicitation} show that the \emph{Practical} prompt performs best for paratextual explicitation, with gains especially pronounced in the thinking mode. We therefore extend the best-performing setting (\emph{Practical} prompt in thinking mode) with agentic retrieval. 

Incorporating bilingual web search yields consistent improvements across all metrics (BLEU: +0.52, ROUGE-L: +3.77, BERTScore F1: +0.52, LLM-as-a-Judge: +2.86). This confirms that access to external resources enables more accurate and contextually-enriched explicitations. The trade-off, however, is efficiency: this retrieval pipeline incurs over ten times the runtime of its non-agentic counterpart.

\subsection{End-to-End Evaluation}
\label{sec:end_to_end_results}

\begin{table*} \small
    \centering
    \begin{tabular}{p{0.15\linewidth} p{0.75\linewidth}}
        \toprule
        \textbf{Term} & \begin{CJK*}{UTF8}{gbsn}泮\end{CJK*} \textbf{(\textit{pàn})} \\
        \midrule
        \textbf{Reference(s)} & \textbf{becoming a \textit{xiucai}}: a scholar who has passed the county level of the imperial civil service examination \citep{sondergard-2008}. \\
        \cmidrule(lr){1-2}
        \textbf{LLM generation} & \textbf{`enter the pond'}: Often associated with scholarly study and the education of young boys. In ancient times, it was common for students to gather by such waters to study and recite classical texts. The term is also used in the context of entering formal education, as students would `enter the pond' to begin their studies at a local academy. \\
        \bottomrule
    \end{tabular}

    \vspace{2ex}

    \begin{tabular}{p{0.15\linewidth} p{0.75\linewidth}}
        \toprule
        \textbf{Term} & \begin{CJK*}{UTF8}{gbsn}解语花\end{CJK*} \textbf{(\textit{jiě yǔ huā})} \\
        \midrule
        \textbf{Reference(s)} & \textbf{`Intelligent Flower'}: The name given by Emperor Xuan Zong of the Tang Dynasty to his pet concubine Yang Guifei, who was known for her affectations \citep{lu-etal-1982}. \\
         & \textbf{`flower that understands speech'}: First used by the Tang emperor Xuanzong in reference to his beloved concubine Yang Yuhuan \citep{mair-and-mair-1989}. \\
        \cmidrule(lr){1-2}
        \textbf{LLM generation} & \textbf{Decoding Flower}: A mythical flower in Chinese folklore, said to have the power to understand and respond to human speech, symbolizing deep communication and emotional connection. \\
        \bottomrule
    \end{tabular}
    
    \caption{Illustrative cases of LLM-generated paratextual explicitations compared to translator ones. The first term (\begin{CJK*}{UTF8}{gbsn}泮\end{CJK*}) shows a contextually accurate but non-canonical explicitation, and the second (\begin{CJK*}{UTF8}{gbsn}解语花\end{CJK*}) a hallucinated one. But these examples raise a key question: how do non-expert readers determine whether generations are hallucinated?}
    \label{fig:llm_examples}

    \vspace{-2ex}
\end{table*}

Having examined culture-bound identification and explicitation independently, we next assess model capacity in an end-to-end setting. For terms not identified, automatic metrics pair translator references with an empty string, resulting in a score of zero for that paratextual explicitation.

Table~\ref{tab:end_to_end} shows that the \emph{Practical} prompt achieves the highest scores, suggesting that while it captures fewer terms overall (shown in Table~\ref{tab:term_identification}), the paratexts it generates are of higher quality. Thinking improves performance further across all prompts, enhancing both identification coverage and explicitation quality.

Taken together, the three evaluation stages reveal the nature of this task: identification benefits from theoretically-grounded criteria for recognizing culture-specific expressions, whereas explicitation is inherently more audience-oriented, favoring prompts that foreground communicative clarity. Reasoning-enabled inference further amplifies both effects, suggesting that structured intermediate processing helps align system behavior with human translation practices. These findings highlight how no single prompting strategy suffices across both tasks and that effective paratextual explicitation requires adapting instructions to meet the demands of identification and explanation alike.

Table~\ref{fig:llm_examples} presents examples of LLM-generated paratexts under the \emph{Practical} setting. The first example is accurate, while the second is hallucinated; we provide their corresponding translator explicitations to facilitate interpretation.

\subsection{Human Evaluation}
\label{section:human_evaluation}

To complement these automated results, we conduct two-stage human evaluation to measure the impact and quality of LLM-generated paratexts. This evaluation builds on previous experiments by examining how paratextual explicitations affect perceived translation quality and their alignment to human judgments.

All LLM outputs were generated using \textsc{Qwen3-235B-A22B}, and three native English speakers with no fluency in Chinese (the target audience of the translation of such a text) served as evaluators. Content was anonymized and presented in randomized order to reduce bias; evaluators indicated their preferred translation or explicitation, or selected `no preference' if both were deemed comparable.

\paragraph{Paratext Impact.}
We first assess whether human evaluators prefer paratexts in LLM translations. Two subsets of stories from the \textit{Liaozhai} dataset were selected: (1) four stories translated by all four human translators, all of whom included paratexts; and (2) five stories translated by two human translators, neither of whom included paratexts. For each story, evaluators compared a baseline LLM translation against a paratext-enriched version, where explicitations were produced using the \emph{Theoretical} prompt for culture-bound term identification and the \emph{Practical} prompt for explicitation. Results shown below in Figure~\ref{fig:paratext_impact}.

\begin{figure}[h]
    \centering
    \includegraphics[width=\linewidth]{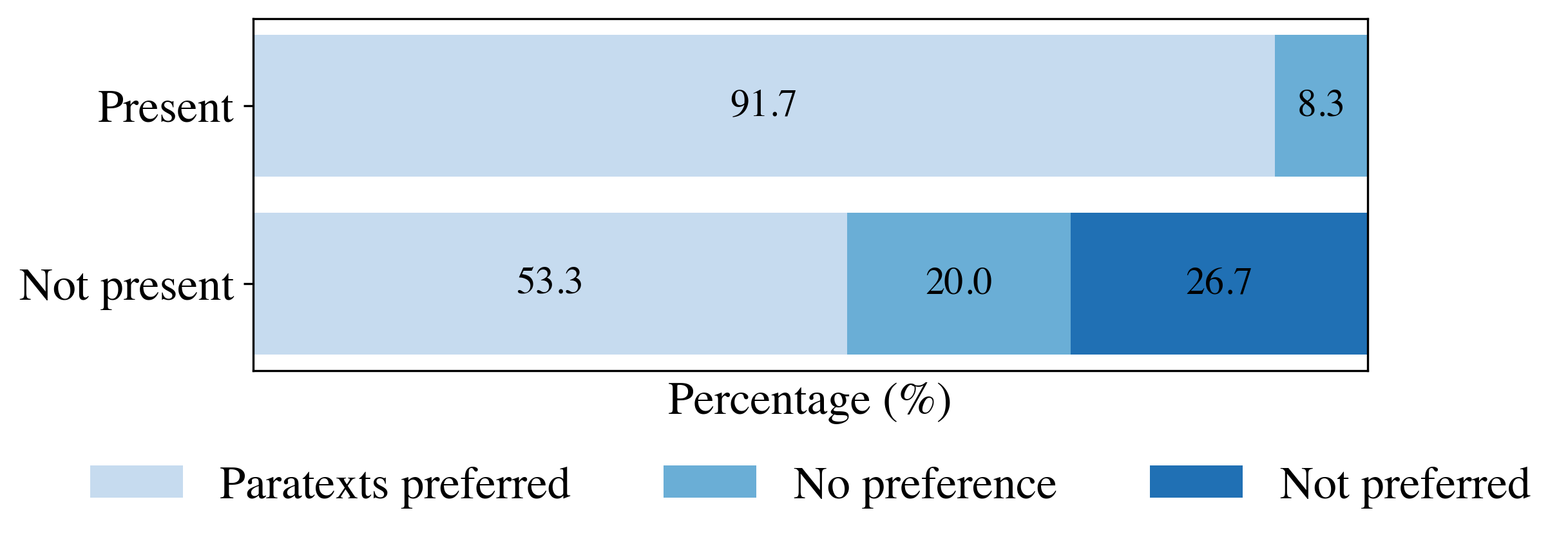}
    \caption{Human preference for LLM translations with versus without paratexts. \texttt{Present}: Stories where all human translators used paratexts. \texttt{Not present}: Stories where all human translators chose not to use paratexts.}
    \label{fig:paratext_impact}
    \vspace{-3ex}
\end{figure}

\paragraph{Explicitation Quality.}
Next, we evaluate the contextual quality of individual explicitations. From the nine \textit{Liaozhai} stories translated by all four translators, we extracted 73 culture-bound terms. For each term, an LLM-generated paratext produced with the \textit{Practical} prompt was compared against one randomly selected translator paratext, both embedded within the full LLM translation. Results shown below in Figure~\ref{fig:explicitation_quality}.

\begin{figure}[h]
    \centering
    \includegraphics[width=\linewidth]{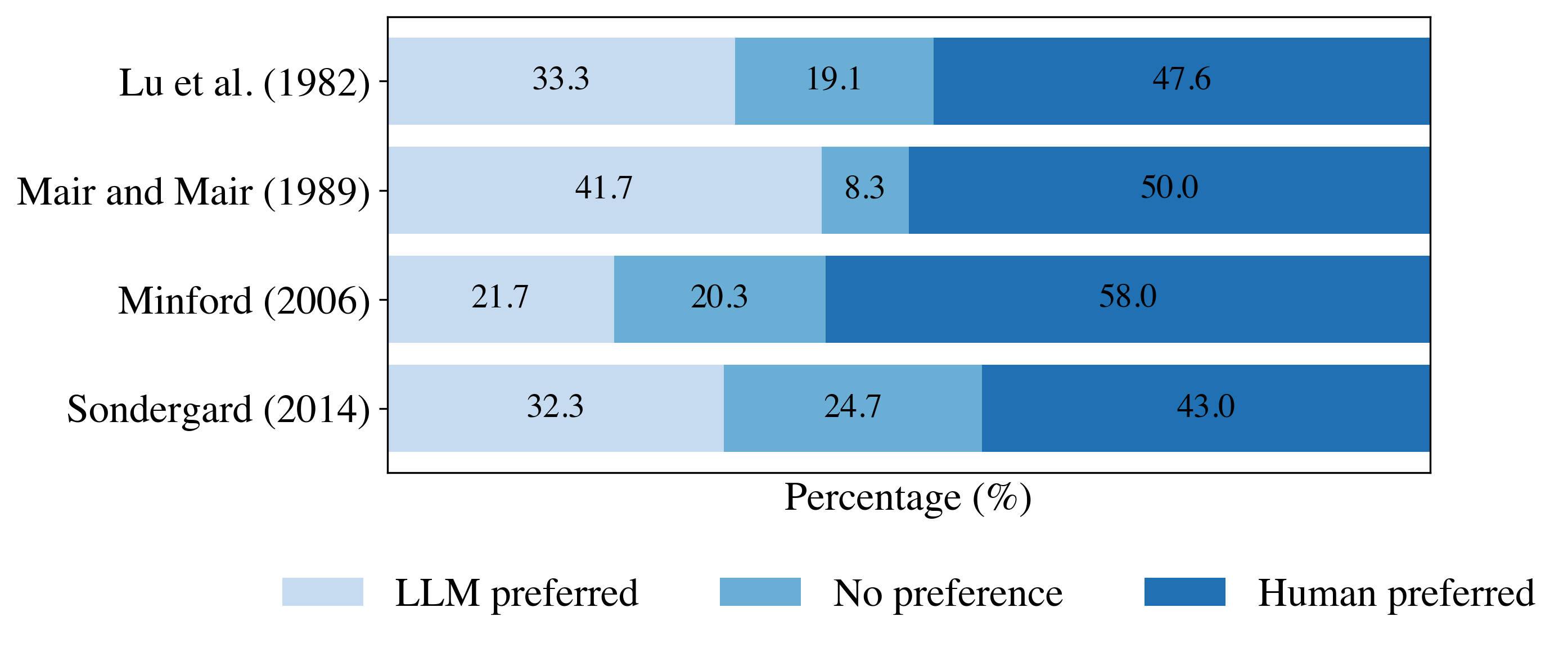}
    \caption{Human preference for LLM versus translator paratexts, reported for each of \citet{lu-etal-1982}, \citet{mair-and-mair-1989}, \citet{minford-2006}, and \citet{sondergard-2008}.}
    \label{fig:explicitation_quality}
    \vspace{-2.5ex}
\end{figure}

\paragraph{Discussion.}
In the paratext impact study, evaluators rated paratext-enriched translations as preferable or equivalent to non paratext-enriched translations in 100\% of stories where human translators included paratexts. Notably, even for stories where no human translators employed paratexts, evaluators still generally preferred the augmented version, suggesting that paratextual explicitations substantially improve clarity for target readers.

In the explicitation quality evaluation, translator paratexts were consistently preferred over LLM-generated ones, indicating that while models can produce plausible explicitations, they still fall short of translator-authored ones. Together, these results highlight the utility of paratexts in improving translation quality as well as the limitations of current LLMs relative to humans.
\section{Analysis}

To better understand the task of paratextual explicitation, we conduct a statistical analysis of patterns in the translator paratexts underlying our dataset. Although translators were not instructed to annotate terms explicitly, each binary choice—whether to provide a paratextual explicitation for a source term—can be treated as a unit of analysis.

\paragraph{Inter-Annotator Agreement.}
To assess consist\-ency across translators, we compute inter-annotator agreement using Krippendorff's Alpha \citep{krippendorff-2011} and pairwise Cohen's Kappa \citep{cohen-1960}. Krippendorff's Alpha provides a reliability coefficient for multiple raters, while Cohen's Kappa estimates agreement between rater pairs beyond chance. The resulting Krippendorff's Alpha of \textbf{-0.3493} points to systematic divergence rather than random variation, a pattern corroborated by pairwise Cohen's Kappa scores ranging from \textbf{-0.45} to \textbf{0.029}. Additional details are provided in Appendix~\ref{appendix:inter-annotator_agreement}.

\paragraph{F1 Scores.}
F1 scores for individual human translators, measured against the pooled set of annotations from the other three, reveal that even experienced translators show limited overlap in term selection (Table~\ref{tab:translator_f1}). This highlights the inherent ambiguity in deciding which elements merit paratextual explicitation and provides a rough estimate of the human upper-bound on our dataset.

\begin{table}[h]
    \centering
    \vspace{0.5ex}
    \begin{tabular}{lr}
        \toprule
        \textbf{Translator} & \textbf{F1-score} \\
        \midrule
        \citet{lu-etal-1982} & 20.74\% \\
        \citet{mair-and-mair-1989} & 24.87\% \\
        \citet{minford-2006} & 29.70\% \\
        \citet{sondergard-2008} & 37.11\% \\
        \bottomrule
    \end{tabular}
    \caption{F1 scores by translator.}
    \label{tab:translator_f1}
    \vspace{-1ex}
\end{table}

\noindent Within this context, our best-performing \emph{Theoretical} identification model achieves an F1 score of 23.73\%—exceeding the `worst-performing' translator, \citet{lu-etal-1982}, by 2.99 points, and trailing the `best-performing' translator, \citet{sondergard-2008}, by 13.38 points. Notably, scores here diverge from the results of human evaluation in \S\ref{section:human_evaluation}, where \citet{minford-2006} is rated the most favorably and \citet{mair-and-mair-1989} the most poorly.

\paragraph{Consensus and Model Performance.}
Not all culture-bound terms are equally important to different translators, and this variation also influences model performance. Terms explicated by multiple translators tend to be more salient, and models likewise find them easier to identify. Table~\ref{tab:identification_consensus} reports the distribution of terms by the number of translators providing explicitations, along with corresponding LLM agreement percentages.

\begin{table}[h]
    \centering
    \begin{tabular}{l|cc}
        \toprule
        \textbf{Explicitated by} & \textbf{Identified} & \textbf{Percentage} \\
        \midrule
        1 translator & 172 / 479 & 35.91\% \\
        2 translators & 36 / 73 & 49.32\% \\
        3 translators & 4 / 5 & 80.00\% \\
        4 translators & 3 / 3 & 100.00\% \\
        \bottomrule
    \end{tabular}
    \caption{Distribution of terms by the number of translators providing paratexts for that term, along with identification accuracy for the best-performing \emph{Theoretical} prompt under the thinking variation. Higher consensus among translators corresponds to higher LLM accuracy.}
    \label{tab:identification_consensus}
    \vspace{-3.25ex}
\end{table}

\paragraph{Variation in Explicitation Content.}
Having established how translators differ in term selection, we next examine how translators vary in paratextual explicitation. Pairwise similarity measures (BLEU and BERTScore) computed over the 81 terms with two or more translator paratexts yield:

\begin{table}[h]
    \centering
    \vspace{-1ex}
    \begin{tabular}{lr}
        \textbf{Bidirectional BLEU:} & 2.03 \\
        \textbf{BERTScore F1:} & 87.40 \\
    \end{tabular}
    \vspace{-2ex}
\end{table}

\noindent Our best-performing LLM setting produces comparable scores (BLEU: 2.14; BERTScore F1: 86.08), suggesting that model-generated paratexts approximate the variability among human translators.

\paragraph{Summary.}
Together, these analyses highlight the considerable interpretive variation in paratextual explicitation. The inherent ambiguity of this task contextualizes our LLM results, showing that even when scores appear low, model outputs often fall within the range of human-human variation.
\section{Related Work}

\paragraph{Cultural and Contextual MT.}
A central challen\-ge in MT is the handling of culture-specific items (CSIs). Conventional adequacy-based metrics often overlook errors on such items, motivating new benchmarks for capturing contextually-appropriate renderings \citep{yao-etal-2024-benchmarking}. In-domain datasets ground this challenge in concrete contexts, such as food translation where lexical choice reflects cultural expectations \citep{zhang-etal-2024-cultural}. Building on these insights, prototype systems extend the focus to end users by detecting cultural references and providing explanations \citep{pandey-etal-2025-culturally}.

\paragraph{Adaptation and Localization.}
Complementary research in cultural MT focuses on strategies for adapting content once CSIs are \mbox{identified}. One line of work localizes named entities to maintain coherence in the target culture \citep{peskov-etal-2021-adapting-entities}. Others enrich translations with additional context, such as through curated explicitation corpora \citep{han-etal-2023-bridging} or by aligning background facts with external knowledge \citep{lou-niehues-2023}. Beyond text, adaptation similarly extends to other mo\-dalities, where image transcreation poses a parallel challenge \citep{khanuja-etal-2024-image}.

\paragraph{Modeling Human Translation.}
Recognizing that professional translators rely on reasoning beyond surface patterns, recent research has sought to computationally model these decision-making processes. This includes retrieval-augmented approaches that incorporate external knowledge during inference \citep{wang-etal-2025-retrieval-augmented}, as well as agent-based methods that simulate collaborative workflows between translators, editors, and proofreaders \citep{wu-etal-2025-perhaps-beyond}.

These directions parallel long-standing concerns in translation studies, where applied research has traditionally emphasized translator training, tools, and quality assessment \citep{holmes-1972}. Translation aids have then been categorized into software tools, reference resources, and collaborative environments \citep{pym-2007}, categories that now find computational analogues within the systems emerging in MT research.
\section{Conclusion}

In this paper, we present a first study of paratextual explicitation in literary MT. Drawing on insights from translation studies, we explore the poetics of paratexts as liminal devices that shape the interaction between a source text and target reader from the borderlands of a work. Methodologically, we formalize this task through the construction of an expert-aligned dataset of classical Chinese stories, evaluate contemporary LLMs on choice and content of explicitation, and analyze variation across professional translators to situate model performance within human practice.

While we focus on literary MT, the relevance of paratextual explicitation extends well beyond such a setting. In monolingual contexts, paratexts can serve as a form of explanatory glossing, clarifying technical or domain-specific terms for non-expert audiences. Similarly, in personalized applications, paratexts can be tuned to a reader's prior knowledge, interests, or expertise. These fine-grained levels of tailoring are impractical for human translators yet may be achievable through computational methods.

Paratextual explicitations thus offer more than just peripheral embellishment. Translations rarely preserve the full cultural, historical, and stylistic fabric of their originals, and paratexts provide a pragmatic means of bridging the contextual gap left by literal renderings. In this spirit, paratextual explicitation functions as a finely calibrated \textit{looking-glass} into adaptation—refracting the interpretive choices, necessary compromises, and subtle negotiations through which texts are continually reshaped for new readers and new worlds.
\section*{Limitations}

\paragraph{Paratext Types.}
Genette's \citeyearpar{genette-1987} framework for paratexts draws a distinction between between peritexts—elements included in the same volume as the main text, such as notes or glossaries—and epitexts, which exist independently but relate to the main text, such as interviews or promotional materials \citep{munday-2016}. Subsequent scholarship has extended this framework to include additional elements encompassing material features of the text (e.g., typeface, binding, page layout) as well as digital artifacts (e.g., metadata, hyperlinks); see \citet{batchelor-2018}. Translation studies has long recognized that paratexts shape how translated works are received and interpreted, and some scholars consider translations themselves as paratexts relative to the original.

This paper focuses specifically on paratextual explicitation through notes and commentary and does not address other forms within or beyond Genette's framework, such as translator prefaces or author bibliographies. These additional materials are included in the released dataset, and we invite future work to explore a broader range of paratextual forms across both textual and visual modalities.

\paragraph{Language and Domain.}
Our dataset exclusively focuses on classical Chinese to English translation within the literary domain. This allows us to study paratextual explicitation in a setting that is linguistically complex and culturally rich, but also constrains the generalizability of our findings. Literary texts present a unique challenge in terms of stylistic variation, cultural references, and interpretive nuance, which may not fully reflect patterns in other genres. Nevertheless, the underlying task of paratextual explicitation is not inherently limited to this language pair or domain and can be applied to a wide range of contexts.

\paragraph{Evaluation.}
Paratextual explicitation involves free-form generation rather than constrained translation, meaning that multiple formulations of the same information can be valid. To assess model output in this setting, we adopted a \mbox{complementary} suite of automated metrics: BLEU \citep{papineni-etal-2002-bleu} to provide a precision-based measurement, ROUGE-L \citep{lin-2004-rouge} to capture longer subsequence matches, and BERTScore \citep{zhang-etal-2020} to evaluate semantic similarity across different surface realizations. While these metrics offer a useful baseline for measuring overlap and meaning, they remain limited in capturing the cultural nuance required by paratextual explicitation.

More recent metrics such as BLEURT \citep{sellam-etal-2020-bleurt} and COMET \citep{rei-etal-2020-comet} aim to evaluate semantic quality and general translation adequacy, yet still fall short of assessing whether paratexts are factually accurate, contextually relevant, useful to the reader, or expressed in an appropriate manner. Human evaluation frameworks such as MQM \citep{lommel-2013-multidimensional} provide hierarchical error annotation, but do not fully reflect the qualitative aspects that make explicitation meaningful. These limitations underscore the need for task-specific evaluation methods capable of assessing the context-sensitive and interpretive nature of translation commentary.
\section*{Ethical Considerations}

This project obtained approval from the University of Edinburgh's Informatics Research Ethics committee, application number 2024/160527.

\paragraph{Fair Use.}
The original classical Chinese source text for \textit{Liaozhai zhiyi} is openly accessible \href{https://liaozhai.5000yan.com/}{online}. In accordance with fair use provisions, we release only the paratextual materials and metadata associated with the four human translations, explicitly excluding the stories themselves. These materials are provided in a transformative manner and intended solely for research purposes, ensuring that the dataset does not infringe upon the rights of the original publishers or translators.

\paragraph{Institutional and Ideological Concerns.}
Beyond questions of fair use, ethical considerations also involve the role of publishing agents in shaping how translations are produced, framed, and received. Translators, editors, and publishers alike make deliberate choices about which paratexts to include and how to present them, shaping the ways in which certain perspectives are amplified and others minimized. While our work explores paratextual explicitation as a computational task, it remains situated within broader institutional and ideological contexts which influence how knowledge and norms are transmitted across cultures. More broadly, discussions of gender, feminism, and queer representation in translation \citep[\textit{inter alia}]{godard-1990, simon-1996, harvey-2012}—though beyond the scope of this work—illustrate how paratextual practices can influence the voices marginalized or silenced in translation.
\section*{Acknowledgments}

This work received funding from the EU’s Horizon Europe (HE) Research and Innovation programme [grant numbers 101070631, 101070350] and from UK Research and Innovation under the UK government’s HE funding guarantee [grant numbers 10039436, 10052546].

\nocite{*}
\bibliography{custom}

\appendix

\section{Exclusion of Giles (1880)}
\label{appendix:giles}

\begin{table*} \small
    \centering
    \begin{tabular}{p{0.1\linewidth} p{0.85\linewidth}}
        \toprule
        \textbf{Subsystem} & \textbf{Paratext} \\
        \midrule
        literary & \textbf{We `know each other's sound'}: One who `knows the sound' of another is, as William Acker puts it: \newline
            \hspace*{2em}
            \begin{minipage}{\dimexpr\linewidth-2em}
                a friend whose knowledge of music is such, and whose mind is so attuned to that of the player that he can catch the finest nuances of the performer's thought and feeling, as he listens, and by his speech or by his silence after the playing of a piece shows that he has understood the other's thoughts as though they have been spoken rather than played...
            \end{minipage}
                \hspace*{12em} (\textit{Some T'ang and Pre-T'ang Texts on Chinese Painting} (Leiden, 1954), p. 10) \newline
        The expression comes from the story of Bo Ya and Zhong Ziqi, in the Taoist \textit{Book of Liezi} \citep{minford-2006}. \\
        \cmidrule(lr){1-2}
        cultural & \textbf{the Cut Sleeve persuasion}: Emperor Ai, last ruler of the Former Han dynasty (206 BC-AD 9), had a number of boy-lovers, the best-known of whom was a certain Dong Xian. Once when the Emperor was sharing his couch with Dong Xian, the latter fell asleep lying across the Emperor's sleeve. When the Emperor was called away to grant an audience, he took his sword and cut off his sleeve rather than disturb the sleep of his favourite. Hence the term `Cut Sleeve' (\textit{duanxiu}) has become a literary expression for homosexuality among men \citep{minford-2006}. \\
        \cmidrule(lr){1-2}
        social & \textbf{The white clothes of the xiucai}: Worn by a scholar who's passed the imperial civil service examination at the county level \citep{sondergard-2008}. \\
        \cmidrule(lr){1-2}
        supplemental & \textbf{A notorious place}: The Bu River in Shandong province passed into the vernacular as a "place notorious for profligacy" since it became a popular site for romantic trysts (see Zhu 51n6) \citep{sondergard-2008}. \\
        \bottomrule
    \end{tabular}
    \caption{Representative examples of paratexts illustrating the literary, cultural, social, and supplementary subsystems. The two paratexts from \citet{minford-2006} are quite extensive and have been condensed here for ease of reading.}
    \label{tab:subsystem_examples}
    \vspace{-2.5ex}
\end{table*}

Herbert A. Giles (1845–1935) was a British diplomat and professor of Chinese at the University of Cambridge, renowned for his extensive translations of classical Chinese literature and scholarly work shaping Western understanding of Chinese language and culture. His \citeyear{giles-1880} translation of \textit{Liaozhai}, \href{https://www.gutenberg.org/ebooks/43629}{\textit{Strange Stories from a Chinese Studio (Volumes 1 and 2)}}, is credited with introducing the literary work to Western audiences.

Giles's translation reflects the linguistic conventions of his era, employing his eponymous Wade-Giles romanization system to render the names of people and places. This style has since been largely supplanted by \textit{pinyin} romanization (e.g., `Peking' in Wade-Giles is `Beijing' in \textit{pinyin}), resulting in a naming convention that differs substantially from modern usage.

In addition, Giles's translation exhibits certain interpretive liberties, particularly with material he considered inappropriate or sensitive. For instance, depictions of fox spirits entering a bedchamber at night are altered to more innocuous events such as drinking tea. While consistent with the conventions of his time, such editorial interventions reflect broader patterns of ideological rewriting in translation.

Today, Giles's translation is best regarded as a historical artifact. Translations reflect the period they were produced in, and modern Western understanding of Chinese culture and literature has evolved considerably since the nineteenth century. Examples provided below.

\vspace{1ex}

On \textit{`bamboo shoots'}, Giles writes:
\begin{quote}
    Which, well cooked, are a very good substitute for asparagus.
\end{quote}

On \textit{the `Silver River'}, Giles writes:
\begin{quote}
    The Milky Way is known to the Chinese under this name—unquestionably a more poetical one than our own.
\end{quote}

On \textit{Mr. Chang}, Giles writes:
\begin{quote}
    The surnames Chang, Wang, and Li, correspond in China to our Brown, Jones, and Robinson.
\end{quote}

\noindent These examples illustrate the dated and interpretive nature of Giles's commentary. For these reasons, we exclude his work from the dataset used in the experiments reported in this paper.

\vspace{3ex}

\section{Characteristics of Other Translations}

\citet{lu-etal-1982} adopt a conservative approach that adheres closely to classical Chinese narrative structures, using paratexts sparingly and only when deemed strictly necessary. By contrast, \citet{mair-and-mair-1989} weave explicitations directly into the main text, replacing potentially obscure terms with more accessible equivalents instead of heavily relying on external commentary.

\citet{minford-2006} aims to preserve narrative flow while providing paratextual support, including detailed explanations through particularly extensive back matter. \citet{sondergard-2008} pursues the most annotation-intensive strategy, employing frequent footnotes to explicate terms likely to be unfamiliar to the contemporary reader.

We present representative paratexts for each translator below.

{
\centering
\small

\begin{quote}
    \textbf{Flower-Morning:} The twelfth day of the second month, traditionally held to be the birthday of flowers.
\end{quote}

\begin{center}
    \textbf{Note:} \citet{mair-and-mair-1989} for story \texttt{v11s2}, \textit{Yellow-Bloom}.
\end{center}

\vspace{1ex}

\begin{minipage}{\dimexpr\linewidth-2em}
    This, one of the best known and most often anthologized and translated of all the \textit{Tales}, is a greatly expanded variation on a brief item in the much earlier collection \textit{In Search of Spirits}, attributed to Gan Bao (\textit{fl}. 320). In the earlier story, the magician is called Xu Guang: \\

    \hspace*{2em}
    \begin{minipage}{\dimexpr\linewidth-2em}
        Once he was performing his magic arts in the marketplace and begged for a gourd from a vendor, who refused to give him one. So he asked for a flower and planted it in the ground, where it immediately started growing, spreading its tendrils over the ground. First it bore flowers, and then fruits. Xu Guang picked one, ate it, and then began handing the fruits out to the spectators. When the vendor turned to look at his own gourds, they had all disappeared. \\
    \end{minipage}
    
    (My translation of the extract quoted by Zhu Yixuan, \textit{Liaozhai zhiyi ziliao huibian}, revised edition (Tianjin, 2002), p. 17. For the complete tale, see Li Qi and Liang Guofu (eds.), \textit{Soushenji Soushen houji yizhu} (Jilin, 1997), p. 27.) \\[-0.5ex]
    
    For obvious reasons this tale has always been popular with Marxist commentators, and is placed first in the popular selection made by Yan Weiqing and Zhu Qikai in 1984. It has been published many times in cartoon-strip form. \\[-0.5ex]
    
    The Chronicler of the Strange appends one of his most trenchant comments to this tale, sharply reproaching the nouveaux riches for their meanness, for the way they turn a deaf ear to needy friends or relations coming to them with simple requests for loans of food or money. In other words, his target is far broader than the country bumpkin who is made to look such a fool in the tale.
\end{minipage}

\vspace{1ex}

\begin{center}
    \textbf{Note:} \citet{minford-2006} for story \texttt{v1s14}, \textit{Growing Pears}.
\end{center}

\newpage

\begin{quote}
    \textbf{\textit{xiaolian}}: An old term for \textit{juren} (\begin{CJK*}{UTF8}{gbsn}举人\end{CJK*}), a successful candidate in the imperial examination at the provincial level in the Ming and Qing Dynasties.
\end{quote}

\begin{center}
    \textbf{Note:} \citet{lu-etal-1982} for story \texttt{v1s6}, \textit{A Wall-painting}.
\end{center}

\begin{quote}
    \textbf{This poem}: This twelve-line poem, consisting of seven characters per line, is structurally reminiscent of the "\textit{jiang shang yin}" ("River Poem") of Li Bo (699-762 C.E.), China's most famous poet. However, its subject and tone are almost precisely the opposite of those in the Li poem, treating Buddhism and nostalgic sadness rather than Daoism and exuberant joy.
\end{quote}

\begin{center}
    \textbf{Note:} \citet{sondergard-2008} for story \texttt{v2s40}, \textit{Fourth Lady Lin}.
\end{center}
}

\vspace{3ex}

\section{Dataset Structure}

Our \href{https://github.com/sherrieshen/liaozhai}{dataset} is organized into three subfolders:

\begin{itemize}
    \item \texttt{annotations/} containing all expert-aligned paratexts and their corresponding annotations (\texttt{annotations.csv}), as well as the log of typographical corrections (\texttt{corrections.md});
    \item \texttt{source/} including the Chinese source texts in JSON format, organized by classical (\texttt{classical/main.json}) and contemporary (\texttt{contemporary/main.json}) styles; and
    \item \texttt{translations/} including the five English translations (\texttt{1880\_giles/}, \texttt{1982\_lu\_etal/}, \texttt{1989\_mair\_and\_mair/}, \texttt{2006\_minford}, and \texttt{2008\_sondergard}) in \texttt{main.json} files.
\end{itemize}

\noindent Each \texttt{main.json} contains entries corresponding to texts in the collection, presented in order of appearance. Possible metadata fields for each entry include:

\begin{itemize}[noitemsep]
    \item \texttt{id}: the global identifier with regards to the Chinese source text;
    \item \texttt{title}: the story title;
    \item \texttt{content}: the story body;
    \item \texttt{commentary}: curator notes on literary significance; and
    \item \texttt{notes}: translator paratexts, in the form of footnotes or endnotes.
\end{itemize}

\noindent Not all fields appear in every file, and some translator folders include additional paratextual materials, such as glossaries or appendices.

\vspace{3ex}

\section{Subsystem Examples}
\label{appendix:subsystem_examples}

Representative examples of paratexts classified according to the five-part framework adapted from polysystem theory are provided in Table \ref{tab:subsystem_examples}, covering the literary, cultural, social, and supplementary subsystems. Historical examples are given in the main text (Figure~\ref{fig:dataset_example}) and are therefore not repeated here.

Paratexts from \citet{minford-2006} are particularly extensive; the excerpts included in the table have been condensed to highlight the core explanatory content while maintaining readability. Full annotations for all paratexts are available in the released dataset.

\newpage
\onecolumn

\section{Prompts for Term Identification}
\label{appendix:prompts_identification}

\tcbset{prompt/.style={
    colback=gray!5,
    colframe=gray!90,
    boxrule=0.5mm,
    arc=2mm,
    left=8pt,
    right=8pt,
    top=6pt,
    bottom=6pt,
    fonttitle=\ttfamily,
    fontupper=\ttfamily
}}

\begin{tcolorbox}[prompt]
    \textbf{Default:} \\
    You are a helpful translation assistant. When provided with a story in classical Chinese, identify key terms that require additional explanation when translated into English. Return these terms as a comma-separated list. \\

    \textbf{Theoretical:} \\
    The terms you identify should be culture-bound terms as defined in translation studies: expressions deeply rooted in the literary, historical, or social context of the source culture. Such terms are often unfamiliar to readers from other cultures and may necessitate explicitation to bridge the gap in understanding. \\

    \textbf{Practical:} \\
    Your target audience is composed of native English speakers with limited knowledge of Chinese culture. The terms you identify for additional explanation should therefore help them understand the story or its setting in a more meaningful manner.
\end{tcolorbox}

\vspace{3ex}

\section{Prompts for Term Explicitation}
\label{appendix:prompts_explicitation}

\begin{tcolorbox}[prompt]
    \textbf{Default:} \\
    You are a helpful translation assistant. Given a classical Chinese story and term from the story, provide (1) an English translation of the term and (2) a clear description of the term's meaning or significance. Format your answer as: \{translated\_term\}: \{description\}. \\

    \textbf{Theoretical:} \\
    Select an appropriate translation strategy (e.g., domestication, foreignization) for the term and let that choice guide your rendering and explanation. Interpret the culture-bound term with respect to its role within the literary, cultural, historical, or social dynamics of the source culture and present your description as a peritext in the Genettean sense—a translator's footnote intended to support the reader's understanding. Do not explain your reasoning; simply provide the term and description. \\

    \textbf{Practical:} \\
    Translate the term for a target audience of native English speakers unfamiliar with Chinese culture. Your description should preserve the term's cultural grounding while remaining clear and accessible. Keep the description concise but informative, offering just enough context to aid reader understanding without being overwhelming. \\

    \textbf{Agentic:} \\
    You are an expert at identifying relevant information. From the provided search results, extract passages that seem the most relevant to defining the classical Chinese term in the given context. Focus on dictionary definitions, explanations, and contextual usage information.
\end{tcolorbox}

\newpage
\twocolumn

\section{Experimental Setup}
\label{appendix:experimental_setup}

We follow \textsc{Qwen3}'s recommended hyperparameter settings of \texttt{temperature} = 0.7, \texttt{top\_p} = 0.8, \texttt{top\_k} = 20, \texttt{min\_p} = 0 for non-thinking mode and \texttt{temperature} = 0.6, \texttt{top\_p} = 0.95, \texttt{top\_k} = 20, \texttt{min\_p} = 0 for thinking mode.

For LLM-as-a-Judge, we use \textsc{Qwen3-235B-A22B} under the non-thinking configuration with same hyperparameter settings. In each evaluation, we present the judge with the classical Chinese source term, the LLM-generated explicitation, and all human reference(s), asking it to evaluate accuracy and clarity on a scale of 0 to 100.

\vspace{3ex}

\section{Human Evaluation Details}

\textbf{Instruction for paratext impact evaluation:}

\begin{figure}[h]
    \begin{mdframed}
        \setlength{\parskip}{0.5em}
        For each evaluation, you will be given two English translations of a classical Chinese story from \textit{Liaozhai}. Select the translation you prefer, or no preference if both are comparable. You may optionally provide a brief justification of your choice.
    \end{mdframed}
\end{figure}

\noindent \textbf{Instruction for explicitation quality evaluation:}

\begin{figure}[h]
    \begin{mdframed}
        \setlength{\parskip}{0.5em}
        Read the given English translation of a classical Chinese story from \textit{Liaozhai}. Throughout the translation you will see terms highlighted, and then two explanations of the term. Select the explanation you prefer, or no preference if both are comparable. You may optionally provide a brief justification of your choice.
    \end{mdframed}
\end{figure}

\noindent Each of the 54 story evaluations was compensated at \$3.00 USD, resulting in a total cost of \$162 USD for human evaluation.

\vspace{3ex}

\section{Inter-Annotator Agreement}
\label{appendix:inter-annotator_agreement}

\begin{figure}[h]
    \centering
    \includegraphics[width=\linewidth]{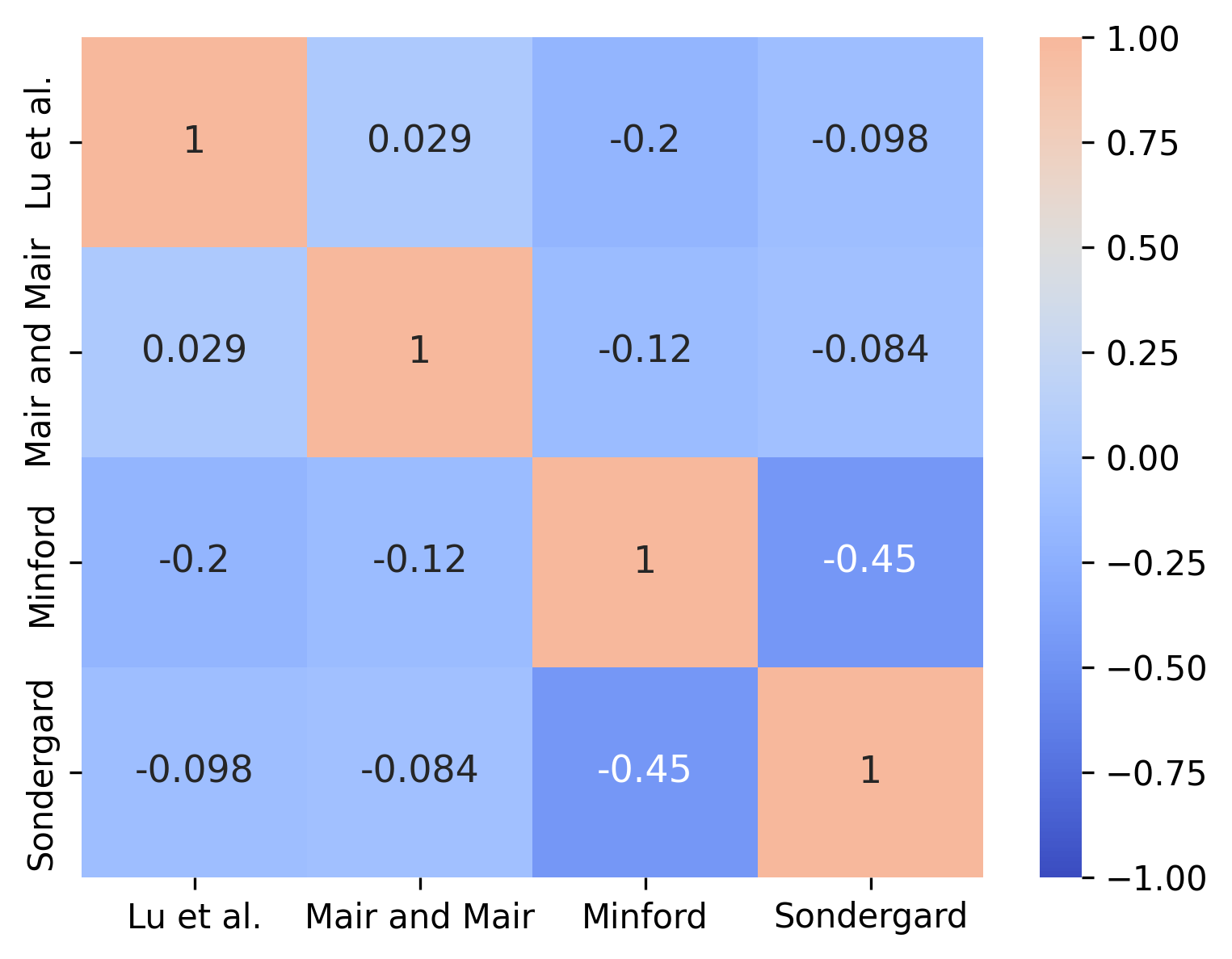}
    \caption{Pairwise Cohen's Kappa heatmap results.}
    \label{fig:pairwise_cohen_kappa}
\end{figure}

For Krippendorff's Alpha, we adopt a three-way encoding scheme for each source term: \texttt{NaN} for stories not translated by a given translator, \texttt{0} for terms left unexplicitated, and \texttt{1} for terms that received explicitation. For Cohen's Kappa, we compute pairwise, chance-corrected agreement between translators on their set of overlapping stories (results in Figure~\ref{fig:pairwise_cohen_kappa}).

This combined approach measures overall agreement among all translators, with pairwise comparisons revealing whether any individual translator disproportionately affects the aggregate score.

\end{document}